\documentclass[10pt,twocolumn,letterpaper]{article}

\usepackage[pagenumbers]{cvpr} 

\usepackage{multirow}

\definecolor{cvprblue}{rgb}{0.21,0.49,0.74}
\usepackage[pagebackref,breaklinks,colorlinks,citecolor=cvprblue]{hyperref}

\def\paperID{xxx} 
\def\confName{xxx\xspace}
\def\confYear{xxx\xspace}

\usepackage{xcolor}
\definecolor{sigmacolor}{RGB}{195,105,55}

\title{CrossDepth: Geometry-Constrained Attention for Generalizable Multi-View Surround Depth Estimation}

\author{Samer Abualhanud\\
Leibniz University Hannover\\
Hannover, Germany\\
{\tt\small abualhanud@ipi.uni-hannover.de}
\and
Max Mehltretter\\
Leibniz University Hannover\\
Hannover, Germany\\
{\tt\small mehltretter@ipi.uni-hannover.de}
}

\begin{document}
\maketitle


\begin{abstract}
Reliable 3D understanding of the surrounding environment is a core requirement for autonomous driving. Multi-view surround camera rigs provide broad scene coverage, but the spatially adjacent images typically overlap only minimally. Consequently, the depth of most pixels must be inferred from monocular appearance cues. These cues can appear differently across images and may therefore be interpreted differently by the depth estimation model. We target two main sources of cross-image inconsistency: differences in camera intrinsics and the limited receptive field of each image. We address the former by conditioning the features on per-pixel camera-aware ray embeddings, enabling the network to account for camera-dependent variations in monocular cues. We address the latter by extending each pixel's context beyond its own image through cross-image attention constrained to geometrically plausible regions, derived from the calibrated rig setup. The model is trained in a fully self-supervised manner based on photometric consistency. Evaluations on DDAD and nuScenes show improved overall depth accuracy and cross-image depth consistency over state-of-the-art self-supervised methods under in-domain and cross-domain evaluation. Code is available at \url{https://abualhanud.github.io/CrossDepthPage/}.

\end{abstract}    
\section{Introduction}
\label{sec:intro}
Many downstream applications in robotics, including path planning, localization and scenario simulation, usually require a 3D map of the environment. Besides relying on direct range sensors such as LiDAR, 3D maps can be reconstructed from one or multiple images, which provide dense and radiometric information at relatively low cost. Given camera intrinsics and known relative poses, estimated per-pixel depth can be back-projected along the corresponding viewing rays to recover the scene geometry in 3D. In this work, we aim to estimate depth from a multi-view surround camera rig.

\begin{figure}[t] 
    \centering
    \includegraphics[width=0.45\textwidth]{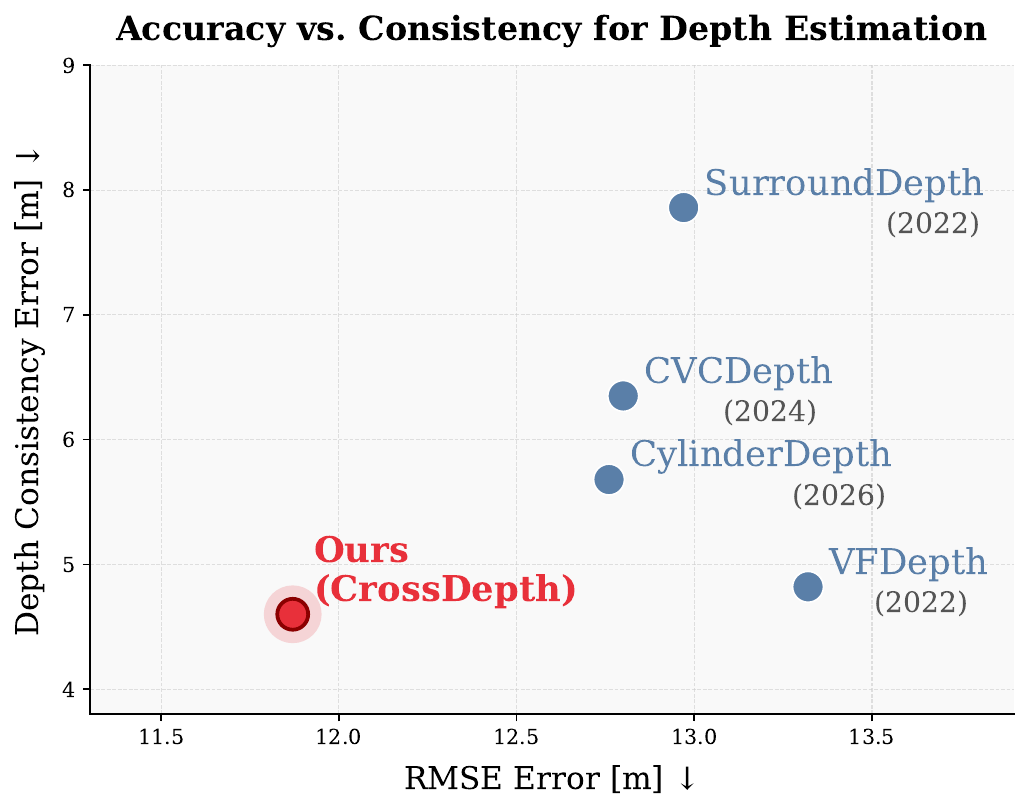}
    \caption{Comparison of our method (red) with recent related methods (blue) in overall depth accuracy, given as RMSE [m], and cross-image depth consistency error [m]. Showing clear improvement in both metrics on DDAD dataset.}
    \label{fig:teaser}
\end{figure}

Surround-view camera systems offer two key advantages. First, their cameras jointly provide complete $360^\circ$ coverage of the environment. Second, the cameras are separated by spatial baselines, resulting in disparity in overlapping regions. This disparity enables self-supervised training of depth estimation through cross-image photometric consistency. When the translation between cameras is given in metric units, it also provides an absolute scale reference, allowing the network to estimate metric depth without ground-truth depth labels. Although images of adjacent cameras overlap, the shared field of view (FOV) is typically limited. Consequently, stereo-based depth estimation is only feasible for few pixels. Depth must therefore mainly be inferred from monocular cues such as object size, semantics, occlusions, perspective, and scene layout. Since the cameras may have different intrinsic parameters, these cues can appear differently across images, causing the same 3D object to have different perspectives and apparent sizes. The model must therefore account for camera-dependent variations in monocular depth cues. Instead of compensating for these variations through focal-length normalization and a global depth scaling factor~\cite{wei2023surrounddepth, Abualhanud_2026_CVPR, ding2024towards, kim2022self}, we condition each pixel on a sinusoidal embedding of its camera ray. This allows the network to learn a non-linear relationship between image appearance, camera intrinsics, and metric depth.

Moreover, monocular cues are embedded in the context of each image. Consequently, the same 3D point observed in multiple images appears under different surrounding contexts, potentially leading to inconsistent depth predictions across images. We therefore apply cross-image attention to exchange contextual information between images and promote cross-image consistency. However, inspired by CylinderDepth~\cite{Abualhanud_2026_CVPR}, we constrain the cross-image attention of each token to a local neighborhood around its expected correspondence in the other images, promoting the aggregation of semantically and geometrically compatible features. To this end, we map token locations from all cameras at a given time step onto a common cylinder.
The cylinder approximates the spatial arrangement of the image planes in the camera rig in which the expected correspondence of each token can be located across all images. We propose reducing the computational cost of both, neighborhood construction and cross-image attention, by discretizing the cylindrical representation. Rather than computing the distance from each token to every other token on the cylinder~\cite{Abualhanud_2026_CVPR}, this discretization allows candidate tokens to be retrieved directly from a local window in the resulting 2D grid. This efficient retrieval enables cross-image attention to be applied to all feature layers passed to the decoder, as opposed to~\cite{Abualhanud_2026_CVPR}, which only applies attention to one coarse resolution.

Because acquiring ground-truth depth is costly and difficult to scale, we train the model fully self-supervised using photometric consistency losses. This removes the dependence on labeled depth data and enables training on more diverse visual domains. 
Thus, our main contributions are:
\begin{itemize}
    \item We propose a novel \textbf{surround-view depth} framework that \textbf{effectively} addresses the main sources of \textbf{cross-image depth inconsistency}.
    \item We propose \textbf{efficient geometry-constrained} \textbf{cross-image attention} over a discretized cylindrical representation. 
    \item We condition image features on per-pixel \textbf{camera-ray embeddings} to account for camera-dependent variations.
    \item We thoroughly evaluate our method on DDAD and nuScenes for \textbf{in-domain} and \textbf{cross-domain} generalization.
\end{itemize}

\section{Related Work}
\label{sec:related}

\paragraph{Monocular Depth Estimation}

Monocular depth estimation methods infer depth from monocular cues. Supervised approaches~\cite{ranftl2021vision, fu2018deep, liu2015learning, eigen2014depth, agarwal2023attention, yang2024depth, yang2024depthv1} learn from ground-truth depth measurements, which are costly to acquire and are often sparse. Self-supervised methods instead derive supervision from the input images themselves by synthesizing a target image from spatially adjacent views~\cite{godard2017unsupervised, garg2016unsupervised}, temporally adjacent frames~\cite{zhou2017unsupervised, godard2019digging, yin2018geonet, guizilini20203d, liu2024mono, watson2021temporal, mahjourian2018unsupervised, ruhkamp2021attention}, or both~\cite{wimbauer2023behind, watson2019self}, and enforcing photometric consistency between the synthesized and real target images. Most existing approaches operate on perspective images with a limited FOV and therefore observe only a fraction of the surroundings. Methods based on omnidirectional images~\cite{vasiljevic2020neural, wang2018self} increase scene coverage by capturing a substantially wider visual context. However, monocular depth estimation from such imagery is more challenging because of the spatially varying distortions, thereby making monocular depth cues less consistent across the image. Moreover, in such single-camera setups, the absence of a spatial metric baseline prevents the model from learning to predict metric depth.


\paragraph{Multi-View Depth Estimation}

When a scene is observed from multiple overlapping images, its geometry can be recovered by matching the visual information shared across the images. Learning-based approaches to this problem broadly follow two paradigms. The first explicitly incorporates projective geometry into the network by warping image features across candidate depth hypotheses and aggregating them in a cost volume, from which a depth map is estimated~\cite{gu2020cascade, im2019dpsnet, yao2018mvsnet, wang2022mvster, khot2019learning, yao2019recurrent}. However, their reconstruction quality generally depends on sufficient overlap between the images. The second, more recent line of work avoids explicit cost-volume construction by directly regressing 3D representations, such as point maps, and relying on attention for cross-image feature sharing~\cite{wang2024dust3r, leroy2024grounding, wang2025vggt, jang2025pow3r, cabon2025must3r, wang2026vggtomega, lin_depth_2026}. However, their reliance on ground-truth depth labels limits scalability, as large amounts of labeled 3D training data are difficult and expensive to obtain. Moreover, these methods are not specifically designed to exploit the known geometry of a calibrated surround-view rig when performing cross-image feature aggregation. In contrast, our method is trained in a self-supervised manner based on photometric consistency and explicitly exploits the rig geometry to constrain feature aggregation across images. \par

\begin{figure*}[ht]
    \centering
    \includegraphics[width=\textwidth]{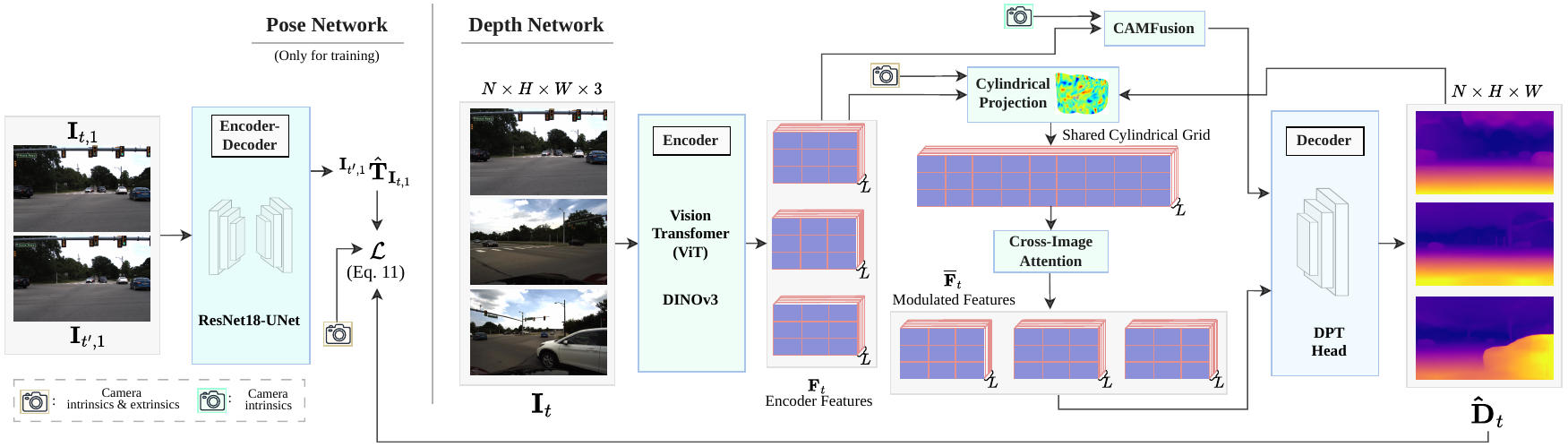}	
    \caption{\normalsize Overview of the proposed method. The depth network independently encodes the target images $\mathbf{I}_{t}$. An initial forward pass predicts depth without cross-image attention, which is then used for cylindrical projection to restrict cross-image attention in a second forward pass, resulting in the final depth prediction $\hat{D}$. The pose network predicts the relative transformation ${}^{\mathbf{I}_{t',1}}\hat{\mathbf{T}}_{\mathbf{I}_{t,1}}$ between the target $\mathbf{I}_{t,1}$ and source front images $\mathbf{I}_{t',1}$.}
    \label{fig:network}
\end{figure*}

\paragraph{Surround Depth Estimation}

Multi-view surround camera rigs provide a wide scene coverage and enable metric depth estimation when the camera baselines are known. Their perspective images also exhibit less distortion than omnidirectional imagery. Depth can be estimated from either a single frame~\cite{guizilini2022full, xu2025self, li2024m2depth, kim2022self, wei2023surrounddepth, ding2024towards, yang2024towards, shi2023ega} or multiple frames~\cite{fei2024driv3r, zou2024m, schmied2023r3d3, yuan2026surroundnexo}. In the single-frame setting, the limited spatial overlap between adjacent views requires depth to be inferred monoscopically for most pixels. Recent work has increasingly investigated self-supervised depth estimation for surround-view camera rigs. FSM~\cite{guizilini2022full} combines spatial and temporal photometric supervision, uses inter-image overlap to recover metric scale, and estimates the individual camera poses between frames. Later approaches~\cite{wei2023surrounddepth, kim2022self} treat the camera system as a rigid rig and predict a single ego-motion. 
CylinderDepth~\cite{Abualhanud_2026_CVPR} incorporates geometric information into cross-image attention by projecting the individual views onto a shared cylinder and weighting attention according to the distance on the cylinder, thereby promoting cross-view consistency. However, CylinderDepth applies cross-image attention only at one coarse feature resolution due to the exhaustive neighborhood search, and its non-learned attention weighting leads to over-smoothed and suboptimal results. In contrast, our method builds on the recent success of foundation models and employs a novel learned cross-image attention that is geometrically constrained and computationally more efficient. In contrast to these methods, we also condition our model on camera intrinsics to account for cross-image depth inconsistencies caused by varying camera intrinsics. 
\section{Methodology}
\label{sec:method}

Given a surround-view camera rig consisting of $N$ time-synchronized cameras with overlapping fields of view, known intrinsic calibration parameters, and metrically scaled relative poses, the objective is to estimate a metric depth map for each input image.
Our depth estimation network follows an encoder-decoder architecture, inspired by~\cite{ranftl2021vision, yang2024depth}. For frame $t$, the input images are denoted by $\mathbf{I}_t \in \mathbb{R}^{N \times H \times W \times 3}$, where $H$ and $W$ denote the image height and width, respectively. Each image $\mathbf{I}_{t,i}$, $i \in N$, is encoded independently of the other images. To this end, it is partitioned into non-overlapping patches and passed through a Vision Transformer (ViT)~\cite{dosovitskiy2020vit}, whose weights are shared across all images. The ViT encoder captures intra-image interactions by applying self-attention at multiple layers over the patches of each image. For the $N$ input images of frame $t$, we denote the collection of features extracted from selected encoder layers by $\mathbf{F}_t = \{\mathbf{F}_{t,l}\}_{l=1}^{L}$, where $\mathbf{F}_{t,l} \in \mathbb{R}^{N \times H_p \times W_p \times F_l}$ denotes the feature maps extracted from encoder layer $l$. Here, $L$ is the number of selected encoder layers whose features are processed by the subsequent modules and passed to the decoder, $H_p$ and $W_p$ denote the spatial dimensions of the patch-level feature grid, and $F_l$ denotes the feature dimension at layer $l$.

The feature maps $\mathbf{F}_{t}$ are processed by our Cross-Image Feature Aggregation module (see Sec.~\ref{sec:attn}), consisting of a projection onto a Shared Cylindrical Grid and Cross-Image Attention.
The decoder is a CNN-based head following the DPT architecture~\cite{ranftl2021vision}, comprising multiple upsampling stages. At each decoder stage, the original features $\mathbf{F}_{t}$ are fused with a Camera-Aware Ray Embedding (see Sec.~\ref{sec:ray}), and combined with the output of the Cross-Image Feature Aggregation module and decoded to predict a depth map for each input image, yielding $\widehat{\mathbf{D}}_t \in \mathbb{R}^{N \times H \times W}$ (see Fig.~\ref{fig:network}).

We train our model in a self-supervised manner using photometric consistency, without ground-truth depth labels. Given the target frame $\mathbf{I}_{t}$, the depth network predicts a depth map for each of the $N$ images and is supervised by spatial photometric consistency across overlapping images. Since this overlap is typically limited, we additionally employ temporal and spatio-temporal supervision, which require the poses of consecutive frames to be known. A pose network takes the front-view images from the target frame $\mathbf{I}_{t}$ and a source frame $\mathbf{I}_{t'}$, where $t' \in {t-1,t+1}$, and predicts their relative temporal transformation. Using the known relative poses of the cameras, this transformation is converted to the corresponding temporal transformation for each camera. These transformations are then used to render the target images from the source frame and enforce temporal and spatio-temporal photometric consistency (see Sec.~\ref{sec:train}).

\subsection{Cross-Image Feature Aggregation}
\label{sec:attn}

Because surround-view rigs have limited image overlap, most depth values must be inferred from monocular cues whose visual context differs across images, leading to inconsistent depth predictions. To enforce cross-image consistency, we apply cross-image attention to the encoder features $\mathbf{F}_t$, expanding the receptive field of each image before passing them to the subsequent stages. Cross-image attention is applied at each extracted encoder layer $l$. However, we constrain attention to geometrically plausible regions. To obtain the geometric constrains, we first perform an initial forward pass without cross-image attention and predict an intermediate depth estimate. Using this depth together with the camera intrinsics and extrinsics, we determine a geometrically plausible neighborhood around the expected correspondence of each token in the other images. During a subsequent second forward pass, the cross-image attention is enabled and restricted to tokens within these neighborhoods. This reduces ambiguity in feature matching by excluding geometrically implausible candidates that may appear similar in feature space. Since intra-image interactions have already been modeled by the encoder’s self-attention layers, we restrict cross-image attention to tokens from other images. This restriction reduces computational cost while promoting cross-image feature aggregation.

\begin{figure}[t]
    \centering

    \includegraphics[width=\linewidth]{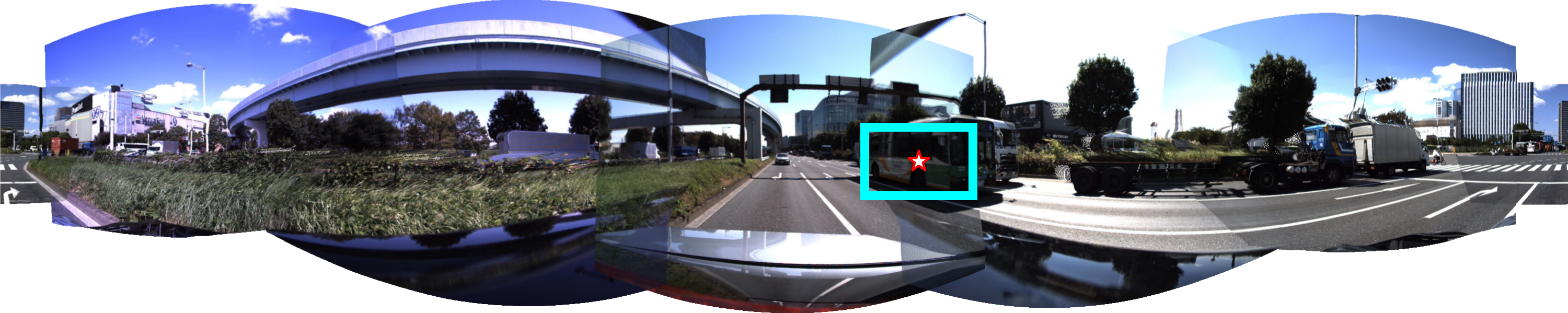}

    \vspace{0.5em}

    \includegraphics[width=0.495\linewidth]{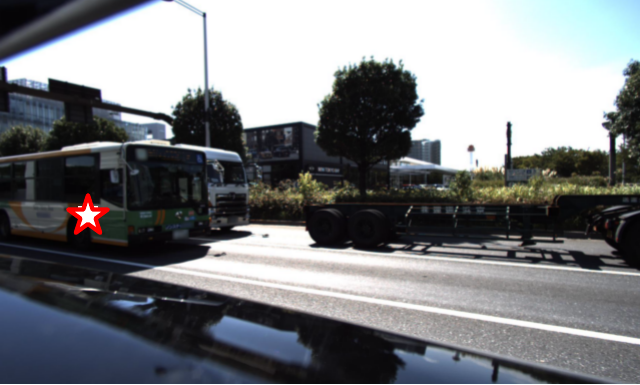}
    \hfill
    \includegraphics[width=0.495\linewidth]{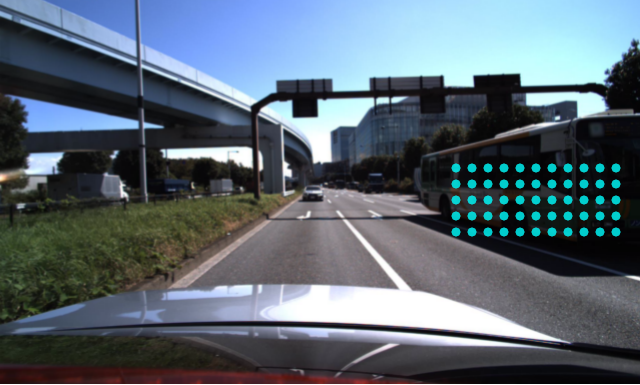}

    \caption{\normalsize Visualization of the flattened cylindrical representation, where overlapping image regions are mapped to nearby locations. For a query token (red star), candidate tokens in the other image are retrieved from a local neighborhood window on the cylinder (blue box) that spans multiple images. Valid cross-image candidates within this window are shown as blue dots in the other overlapping image.}
    \label{fig:attn}
\end{figure}

\paragraph{Shared Cylindrical Grid}

To restrict cross-image attention to geometrically plausible candidates, we first map tokens from all images into a shared representation. We use a vertical cylindrical representation, inspired by CylinderDepth~\cite{Abualhanud_2026_CVPR}, because it is well aligned with the commonly employed horizontal $360^\circ$ arrangement of surround-view camera rigs. Each token is mapped onto the shared cylinder. As a result, overlapping regions from the individual images are mapped to nearby locations on the cylinder, enabling neighborhood retrieval that spans multiple images (see Fig.~\ref{fig:attn}).

To construct the shared cylindrical representation, each feature token $s \in \{1,\ldots,Z\}$, where $Z = N H_p W_p$, is mapped to 3D and expressed w.r.t a common ego coordinate system using the preliminary depth, camera intrinsics, and camera-to-ego transformations. The origin of the ego coordinate system is defined at the centroid of the camera projection centers. The resulting 3D point is mapped onto the shared cylinder and represented by its angular and vertical coordinates, denoted by $(\theta_s, h_s)$, where $\theta_s \in [-\pi,\pi]$. To determine the neighborhood of each token, we identify tokens that lie nearby on the shared cylinder. Because tokens from all images are projected into the same cylindrical space, the resulting neighborhoods can span multiple images. However, rather than performing an exhaustive search by computing pairwise distances between all tokens to find the neighborhood of each token, as in~\cite{Abualhanud_2026_CVPR}, we discretize the cylindrical space into a 2D grid that enables efficient neighborhood retrieval. The cylindrical surface is discretized into $B_{\theta}$ angular bins and $B_h$ vertical bins. The grid coordinates of token $s$ are computed as:
\begin{equation}
u_s =
\left\lfloor
B_{\theta}\frac{\theta_s+\pi}{2\pi}
\right\rfloor,
\quad
v_s =
\left\lfloor
B_h
\frac{h_s-h_{\min}}
{h_{\max}-h_{\min}}
\right\rfloor,
\label{eq:cylindrical_grid}
\end{equation}where $h_{\max}$ and $h_{\min}$ denote the maximum and minimum height values of the vertical cylindrical coordinates, respectively. For each query token $s$, candidate key tokens are gathered from an $K_W \times K_W$ window centered at $(u_s,v_s)$. The horizontal offset is $\Delta u \in \left[-\frac{K_W-1}{2}, \frac{K_W-1}{2}\right]$, and the corresponding angular cell index is given by $\widetilde{u}_s = (u_s+\Delta u) \bmod B_{\theta}$. Thus, windows that cross an angular boundary wrap around to the opposite side of the grid. Similarly, the vertical offset is $\Delta v \in \left[-\frac{K_W-1}{2} ; \frac{K_W-1}{2}\right]$, and the corresponding vertical cell index is given by $\widetilde{v}_s = v_s+\Delta v$. Because the vertical dimension is not periodic, cells outside its valid range are ignored. As the number of tokens may be higher than the number of cells in the cylindrical grid, each grid cell can store more than one token. Therefore, each query's window contains up to $M={K}^2_W \cdot K_{\text{max}}$ candidate key tokens, where $K_{\text{max}}$ denotes the maximum number of tokens that can occupy a single grid cell, which is dynamically determined from the cylindrical projection as the maximum number of tokens assigned to any grid cell over the entire grid for each input frame.

\paragraph{Cross-Image Attention}
The encoder feature maps of all $N$ images at layer $l$, denoted by $\mathbf{F}_{t,l}$, are flattened into $\mathbf{\widetilde{F}}_{t,l} \in \mathbb{R}^{T \times F_l}$. For multi-head attention with $N_h$ heads of dimension $C_h$, the query, key, and value representations for each head $h$ are obtained as $\mathbf{Q}^{(h)} = \mathbf{\widetilde{F}}_{t,l}\mathbf{W}^{(h)}_Q$, $\mathbf{K}^{(h)} = \mathbf{\widetilde{F}}_{t,l}\mathbf{W}^{(h)}_K$, and $\mathbf{V}^{(h)} = \mathbf{\widetilde{F}}_{t,l}\mathbf{W}^{(h)}_V$, respectively, where $\mathbf{W}^{(h)}_Q, \mathbf{W}^{(h)}_K, \mathbf{W}^{(h)}_V \in \mathbb{R}^{F_l \times C_h}$ are learned projection matrices. For each query token $s$, let $m \in \{1,\ldots,M\}$ index one of its $M$ candidate keys. The similarity between query token $s$ and its $m$-th candidate key token is computed as the scaled dot product $S^{(h)}_{s, m} = {\mathbf{Q}^{(h)}_{s}}^{\top}\mathbf{K}^{(h)}_{m}/\sqrt{C_h}$, where $\mathbf{Q}_s^{(h)} \in \mathbb{R}^{C_h}$ is the query vector of token $s$, and $\mathbf{K}^{(h)}_{m} \in \mathbb{R}^{C_h}$ is the key vector of its $m$-th candidate. The attention weights are obtained by applying softmax over the $M$
candidate tokens,

\begin{equation}
A_{s,m}^{(h)}
=
\frac{
\exp(S^{(h)}_{s,m})
}{
\sum_{j=1}^{M}\exp(S^{(h)}_{s,j})
}.
\end{equation}

The resulting cross-image features for query token $s$ are then

\begin{equation}
\mathbf{Y}^{(h)}_s
=
\sum_{m=1}^{M}
A^{(h)}_{s,m}\mathbf{V}^{(h)}_{s,m}.
\label{eq:softmax_attention}
\end{equation}
For each of the $Z$ query tokens, the outputs $\mathbf{Y}^{(h)}_s \in \mathbb{R}^{C_h}$ of the $N_h$ attention heads are concatenated into a vector of dimension $N_h C_h$ and projected using a learned matrix
$\mathbf{W}_O \in \mathbb{R}^{N_h C_h \times F_l}$.
The resulting features for all $Z$ tokens are then reshaped to obtain
$\overline{\mathbf{F}}_{t,l} \in \mathbb{R}^{N \times H_p \times W_p \times F_l}$. Applying these operations to all $L$ layers yields $\overline{\mathbf{F}}_t
=
\{\overline{\mathbf{F}}_{t,l}\}_{l=1}^{L}$. Since each of the $Z$ query tokens attends to only $M$ candidate keys rather than all $Z$ tokens, the number of query--key interactions in our attention module is reduced from $Z^2$ to $ZM$, where $M \ll Z$.

\subsection{Camera-Aware Ray Embedding}
\label{sec:ray}

We incorporate a ray embedding for each token before passing them to the decoder. This allows the decoder to interpret monocular cues w.r.t the corresponding camera intrinsics. For example, the same object may appear at different scales under different focal lengths but should still be predicted at the same metric depth. Conditioning on the camera ray enables the model to learn this dependency between visual appearance and camera parameters, reducing camera-induced variation in the predicted depth and improving the consistency of depth predictions across the images. 

The encoder features $\mathbf{F}_{t}$ are concatenated with the camera-aware ray embeddings $\mathbf{R}_{t}$, and fused using the CAMFusion module, which applies a $3\times3$ convolution to the concatenated features:

\begin{equation}
\tilde{\mathbf{F}}_{t}
=
\mathrm{CAMFusion}
\left(
\mathbf{F}_{t} \Vert \mathbf{R}_{t}
\right),
\label{eq:ray_feature_fusions}
\end{equation} where \(\Vert\) denotes
concatenation. For the encoder features $\mathbf{F}_t$, we construct ray-direction maps $\mathbf{B}_t \in \mathbb{R}^{N \times H_p \times W_p \times 3}$, where each token location is associated with the ray direction determined by the intrinsics of its corresponding camera. Following NeRF~\cite{mildenhall2021nerf}, we apply a sinusoidal encoding \({\gamma_L}_{dir}(\cdot)\) with $L_{\mathrm{dir}}$ frequency bands to each ray direction in $\mathbf{B}_t$ to obtain the ray-embedding maps $\mathbf{R}_t$:

\begin{equation}
\mathbf{R}_{t}
=
\gamma_{L_{\mathrm{dir}}}\!\left(\mathbf{B}_{t}\right)
\,
.
\label{eq:ray_embedding}
\end{equation}
We do not encode relative camera poses in the ray embeddings, as they are already incorporated through the geometric constraints of the cross-image attention.

Finally, the DPT decoder progressively upsamples the cross-image modulated features $\overline{\mathbf{F}}_{t}$ and fuses them with the corresponding camera-aware features $\tilde{\mathbf{F}}_{t}$, following the concept of~\cite{ranftl2021vision}, to predict a depth map for each input image: 

\begin{equation}
\hat{\mathbf{D}}_t
=
\text{DPT}(\overline{\mathbf{F}}_{t}, \tilde{\mathbf{F}}_{t})
\label{eq:head}
\end{equation}

\subsection{Self-Supervised Training}
\label{sec:train}

We train our model in a self-supervised manner by enforcing photometric consistency between a target image $\mathbf{I}_{t,i}$ and its rendered image $\hat{\mathbf{I}}_{t,i}$ from a source view. Following~\cite{godard2017unsupervised}, the photometric loss is defined as:
\begin{align}
\mathcal{L}_{\mathrm{photo}}
=
\frac{1}{|\Omega|}
\sum_{\mathbf{p}\in\Omega}
\Bigg[
&\alpha
\frac{1-\mathrm{SSIM}\!\left(
\hat{\mathbf{I}}_{t,i}(\mathbf{p}),
\mathbf{I}_{t,i}(\mathbf{p})
\right)}{2}
\nonumber\\
&+
(1-\alpha)
\left\|
\hat{\mathbf{I}}_{t,i}(\mathbf{p})
-
\mathbf{I}_{t,i}(\mathbf{p})
\right\|_1
\Bigg].
\label{Eq:photo}
\end{align}
where $\alpha=0.85$, $\Omega$ denotes the set of valid pixels, and SSIM~\cite{wang2004image} is the structural similarity,

We consider three photometric supervision settings: spatial, temporal, and spatio-temporal supervision. For spatial supervision, each pixel position $\mathbf{p}_{\mathbf{I}_{t,i}}$ in the target image $\mathbf{I}_{t,i}$ is projected into a spatially adjacent source image $\mathbf{I}_{t,j}$ using the predicted depth $\hat{\mathbf{D}}_{\mathbf{I}_{t,i}}$, the known metric relative pose ${}^{\mathbf{I}_{t,j}}\mathbf{T}_{\mathbf{I}_{t,i}}$ and camera intrinsics $\mathbf{K}_{}^{\mathrm{int}}$:
\begin{align}
\hat{\mathbf{p}}_{\mathbf{I}_{t,j}}
=
\mathbf{K}_{\mathbf{I}_{t,j}}^{\mathrm{int}}\,
{}^{\mathbf{I}_{t,j}}\mathbf{T}_{\mathbf{I}_{t,i}}\,
\hat{\mathbf{D}}_{\mathbf{I}_{t,i}}\,
\left(\mathbf{K}_{\mathbf{I}_{t,i}}^{\mathrm{int}}\right)^{-1}
\mathbf{p}_{\mathbf{I}_{t,i}}.
\label{Eq:warp}
\end{align}
The rendered target image $\hat{\mathbf{I}}_{t,i}$ is then obtained by sampling the source image at the projected pixel locations. This provides metric-scale supervision in regions with spatial overlap.

For temporal supervision, $\mathbf{I}_{t,i}$ is rendered from the temporally adjacent image $\mathbf{I}_{t',i}$. Following~\cite{ding2024towards}, we predict only the temporal pose of the front camera,
${}^{\mathbf{I}_{t',1}}\hat{\mathbf{T}}_{\mathbf{I}_{t,1}}$, and exploit the rigid camera configuration to obtain the temporal pose of a camera image $i$ as ${}^{\mathbf{I}_{t',i}}\hat{\mathbf{T}}_{\mathbf{I}_{t,i}}
=
{}^{\mathbf{I}_{t,1}}\mathbf{T}_{\mathbf{I}_{t,i}}^{-1}
{}^{\mathbf{I}_{t',1}}\hat{\mathbf{T}}_{\mathbf{I}_{t,1}}
{}^{\mathbf{I}_{t,1}}\mathbf{T}_{\mathbf{I}_{t,i}}$. Using this transformation, a target pixel $\mathbf{p}_{\mathbf{I}_{t,i}}$ is projected into the temporal source image $\mathbf{I}_{t',i}$ as: 
\begin{align} \hat{\mathbf{p}}_{\mathbf{I}_{t',i}} = \mathbf{K}_{\mathbf{I}_{t',i}}^{\mathrm{int}}\, {}^{\mathbf{I}_{t',i}}\hat{\mathbf{T}}_{\mathbf{I}_{t,i}}\, \hat{\mathbf{D}}_{\mathbf{I}_{t,i}}\, \left(\mathbf{K}_{\mathbf{I}_{t,i}}^{\mathrm{int}}\right)^{-1}
\mathbf{p}_{\mathbf{I}_{t,i}}. 
\label{Eq:warp_temp} 
\end{align}

For spatio-temporal supervision~\cite{guizilini2022full}, $\mathbf{I}_{t,i}$ is rendered from an image $\mathbf{I}_{t',j}$ captured by a different camera in a different frame. The corresponding transformation is $
{}^{\mathbf{I}_{t',j}}\hat{\mathbf{T}}_{\mathbf{I}_{t,i}}
=
{}^{\mathbf{I}_{t',j}}\hat{\mathbf{T}}_{\mathbf{I}_{t,j}}
{}^{\mathbf{I}_{t,j}}\mathbf{T}_{\mathbf{I}_{t,i}}
$.
Using this transformation, a target pixel $\mathbf{p}_{\mathbf{I}_{t,i}}$ is projected into the spatio-temporal source image $\mathbf{I}_{t',j}$ as:
\begin{align} \hat{\mathbf{p}}_{\mathbf{I}_{t',j}} = \mathbf{K}_{\mathbf{I}_{t',j}}^{\mathrm{int}}\, {}^{\mathbf{I}_{t',j}}\hat{\mathbf{T}}_{\mathbf{I}_{t,i}}\, \hat{\mathbf{D}}_{\mathbf{I}_{t,i}}\, \left(\mathbf{K}_{\mathbf{I}_{t,i}}^{\mathrm{int}}\right)^{-1}
\mathbf{p}_{\mathbf{I}_{t,i}}. 
\label{Eq:warp_spt} 
\end{align}

Applying Eq.~\ref{Eq:photo} to the temporal, spatial, and spatio-temporal rendered target images yields
$\mathcal{L}_{\mathrm{photo,temp}}$,
$\mathcal{L}_{\mathrm{photo,sp}}$, and
$\mathcal{L}_{\mathrm{photo,spt}}$, respectively. The complete training objective combines these photometric terms with an edge-aware depth smoothness loss $\mathcal{L}_{\mathrm{sm}}$~\cite{godard2017unsupervised}, dense depth consistency loss $\mathcal{L}_{\mathrm{DCCL}}$~\cite{ding2024towards}, and multi-view reconstruction consistency loss $\mathcal{L}_{\mathrm{MVRCL}}$~\cite{ding2024towards}:
\begin{align}
\mathcal{L}
={}&
\mathcal{L}_{\mathrm{photo,temp}}
+\lambda_{\mathrm{sp}}\mathcal{L}_{\mathrm{photo,sp}}
+\lambda_{\mathrm{spt}}\mathcal{L}_{\mathrm{photo,spt}}
\nonumber\\
&+\lambda_{\mathrm{sm}}\mathcal{L}_{\mathrm{sm}}
+
\lambda_{\mathrm{DCCL}}\mathcal{L}_{\mathrm{DCCL}}
+
\lambda_{\mathrm{MVRCL}}\mathcal{L}_{\mathrm{MVRCL}},
\label{Eq:loss}
\end{align}
where the $\lambda$ terms weight the individual loss components.

\begin{table*}[t]
\centering
\setlength{\tabcolsep}{5pt}
\renewcommand{\arraystretch}{0.9}
\scriptsize

\resizebox{0.9\textwidth}{!}{%
\begin{tabular}{@{}llcccc@{\hspace{5pt}}cc@{}}
\toprule
& &
\multicolumn{4}{c}{\textbf{Overall} (ID $|$ CD)} &
\multicolumn{2}{c}{\textbf{Overlap} (ID $|$ CD)} \\
\cmidrule(lr){3-6}
\cmidrule(lr){7-8}
Dataset & Method &
Abs Rel &
Sq Rel [m] &
RMSE [m] &
$\delta < 1.25$ [\%] &
Abs Rel &
Depth Cons [m] \\
\midrule

\multirow{6}{*}{\textbf{DDAD}}
& FSM
& 0.201 $|$ - & - $|$ - & - $|$ - & - $|$ -
& - $|$ - & - $|$ - \\

& FSM*
& 0.228 $|$ - & 4.409 $|$ - & 13.43 $|$ - & 68.7 $|$ -
& - $|$ - & - $|$ - \\

& VFDepth
& 0.218 $|$ 0.294 & 3.660 $|$ 3.817 & 13.32 $|$ 10.78 & 67.4 $|$ 45.1
& 0.222 $|$ 0.332 & \underline{4.82} $|$ 7.27 \\

& SurroundDepth
& 0.208 $|$ 0.344 & 3.371 $|$ 4.305 & 12.97  $|$  11.77 & 69.3 $|$ 36.1
& 0.217 $|$ 0.368 & 7.86 $|$ 7.18 \\

& CVCDepth
& \underline{0.203} $|$ 0.266 & \underline{3.363} $|$ 2.912 & 12.80 $|$ 9.33 & 70.6 $|$ 56.6
& \underline{0.204} $|$ 0.279 & 5.89 $|$ 9.06 \\

& CylinderDepth
& 0.207 $|$ 0.258 & 3.503 $|$ \underline{2.835} & \underline{12.76} $|$ 8.81 & \underline{70.9} $|$ \underline{59.5}
& 0.207 $|$ \underline{0.263} & 5.68 $|$ \underline{7.09} \\

& \textbf{CrossDepth (ours)}
& \textbf{0.185} $|$ \textbf{0.224} & \textbf{2.894} $|$ \textbf{2.151} & \textbf{11.87} $|$ \textbf{7.84} & \textbf{74.7} $|$ \textbf{66.0}
& \textbf{0.186} $|$ \textbf{0.214} & \textbf{4.60} $|$ \textbf{4.51} \\
\midrule

\multirow{6}{*}{\textbf{nuScenes}}
& FSM
& 0.297 $|$ - & - $|$ - & - $|$ - & - $|$ -
& - $|$ - & - $|$ - \\

& FSM*
& 0.319 $|$ - & 7.534 $|$ - & 7.86 $|$ - & 71.6 $|$ -
& - $|$ - & - $|$ - \\

& VFDepth
& 0.289 $|$ 0.380 & 5.718 $|$ 6.585 & 7.55 $|$ 8.70 & 70.9 $|$ \underline{57.3}
& 0.277 $|$ 0.383 & 3.57 $|$ \textbf{4.55} \\

& SurroundDepth
& 0.280 $|$ 0.410 & \textbf{4.401} $|$ 5.879 & 7.46 $|$ 8.98 & 66.1 $|$ 50.0
& 0.295 $|$ 0.470 & 6.33 $|$ 9.44 \\

& CVCDepth
& 0.246 $|$ \underline{0.361} & \underline{4.440} $|$ \textbf{4.768} & \underline{6.78} $|$ \underline{8.54} & 76.4 $|$ 44.6
& 0.219 $|$ \underline{0.357} & 3.46 $|$ 5.57 \\

& CylinderDepth
& \underline{0.244} $|$ 0.376 & 6.025 $|$ 5.804 & 6.82 $|$ 8.90 & \underline{80.5} $|$ 46.6
& \underline{0.218} $|$ 0.366 & \underline{2.69} $|$ 5.46 \\

& \textbf{CrossDepth (ours)}
& \textbf{0.238} $|$ \textbf{0.269} & 5.952 $|$ \underline{5.434} & \textbf{6.60} $|$ \textbf{7.60} &  \textbf{81.2} $|$ \textbf{68.8}
&  \textbf{0.216} $|$ \textbf{0.266} & \textbf{2.56} $|$ \underline{4.73} \\

\bottomrule
\end{tabular}%
}

\caption{
Comparison on DDAD and nuScenes under in-domain (ID) and cross-domain (CD) evaluation, reported for the full image and overlapping regions. Each entry is shown as ID $|$ CD. For DDAD, CD denotes training on nuScenes and testing on DDAD; for nuScenes, CD denotes training on DDAD and testing on nuScenes. FSM* denotes results reproduced by~\cite{kim2022self}. Best results are shown in bold and second-best are underlined.
}
\label{tab:method_compar}
\end{table*}

\begin{figure*}[t]
\centering
\setlength{\tabcolsep}{1pt}
\renewcommand{\arraystretch}{0.0}
\begin{tabular}{*{6}{c}}
\scriptsize Image &
\scriptsize CrossDepth (Ours)&
\scriptsize CylinderDepth&
\scriptsize CVCDepth&
\scriptsize SurroundDepth &
\scriptsize VFDepth\\

\includegraphics[width=0.16\textwidth]{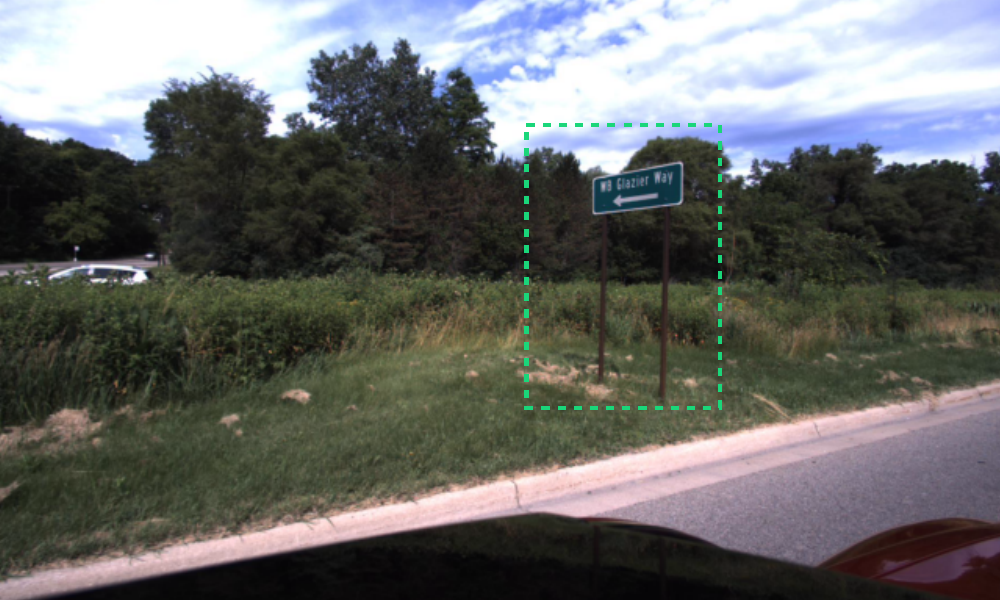} &
\includegraphics[width=0.16\textwidth]{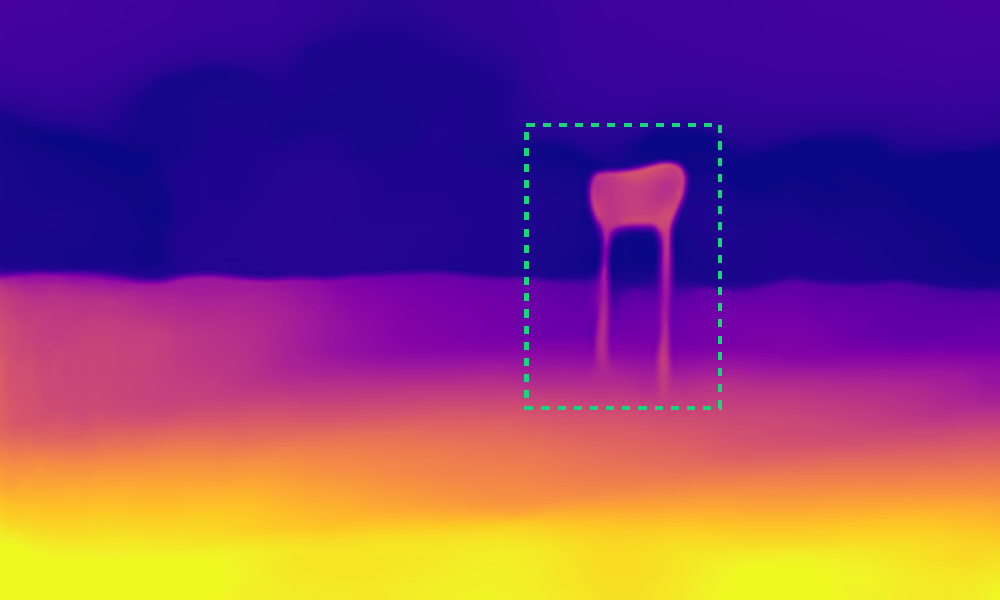} &
\includegraphics[width=0.16\textwidth]{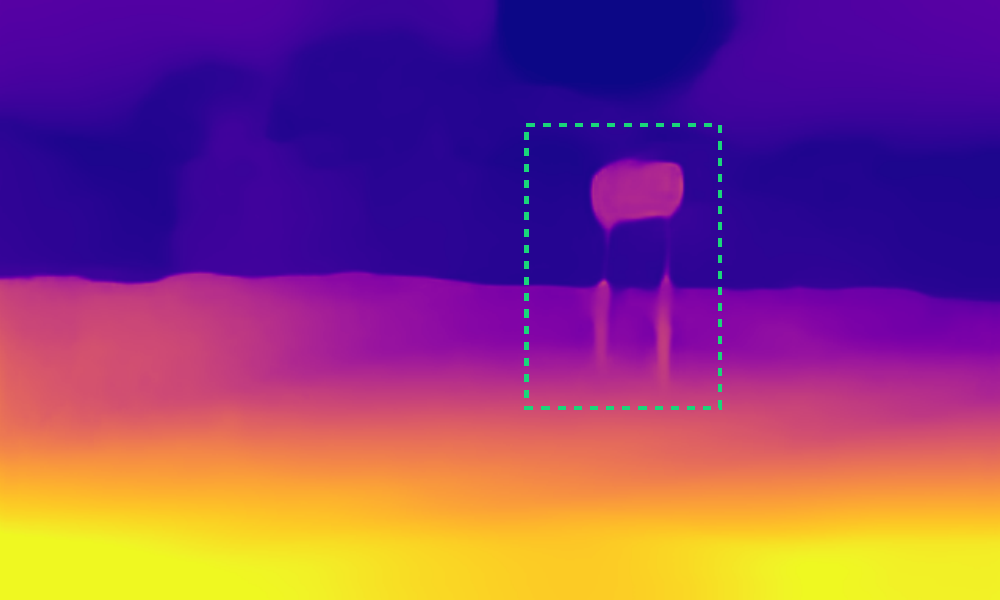}&
\includegraphics[width=0.16\textwidth]{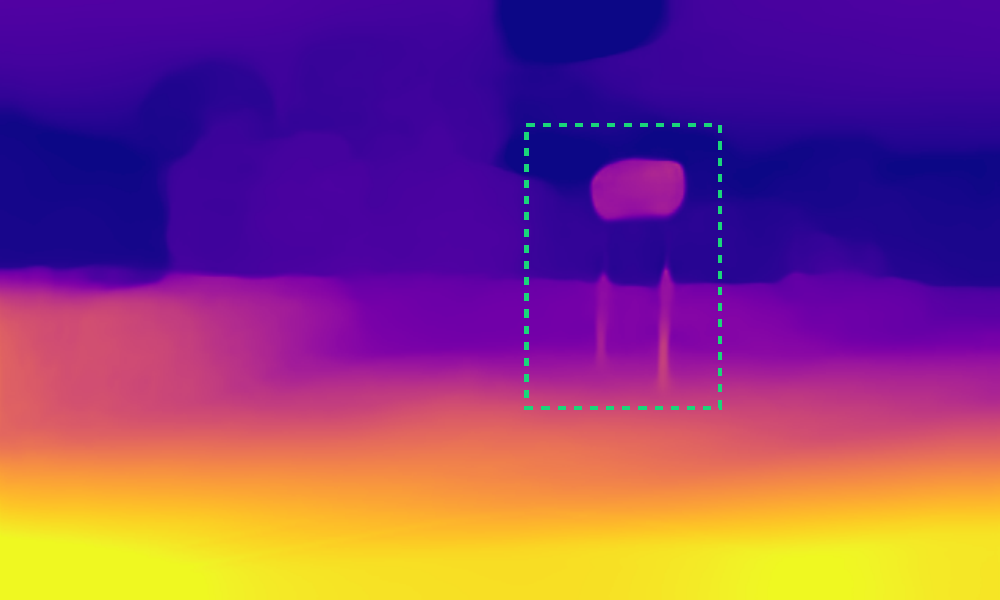}&
\includegraphics[width=0.16\textwidth]{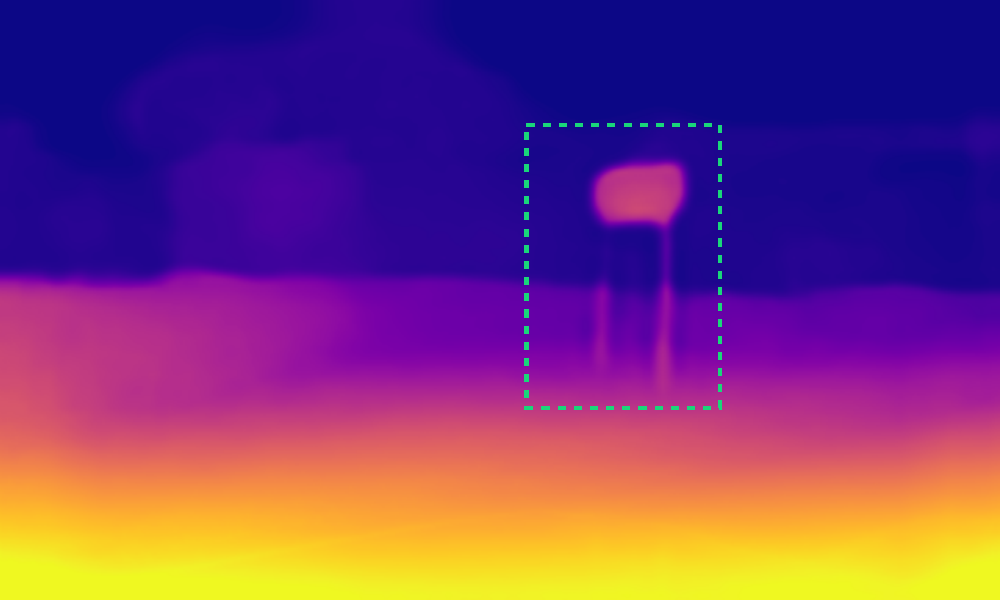} &
\includegraphics[width=0.16\textwidth]{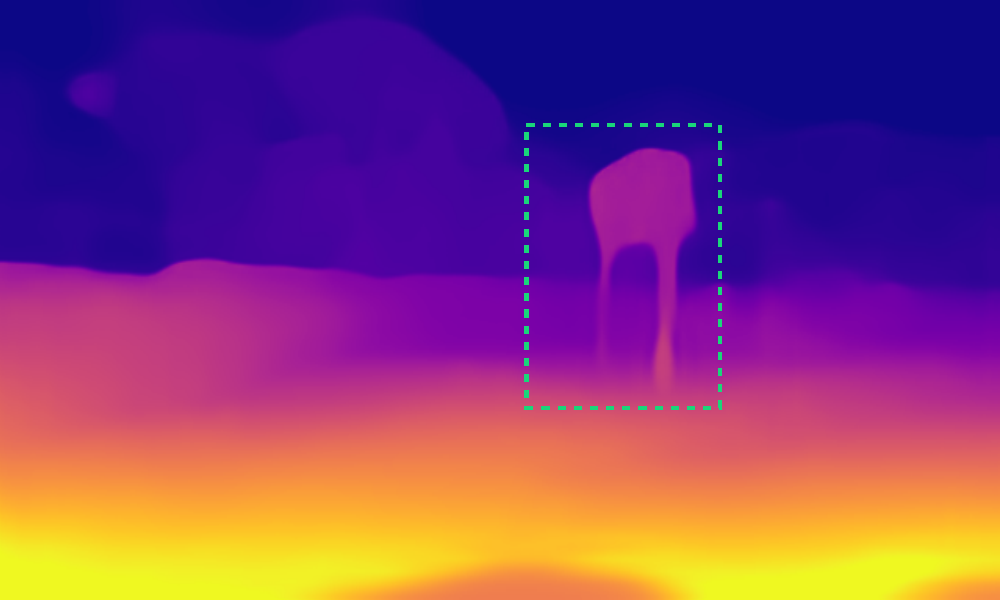} \\[2pt]

\includegraphics[width=0.16\textwidth]{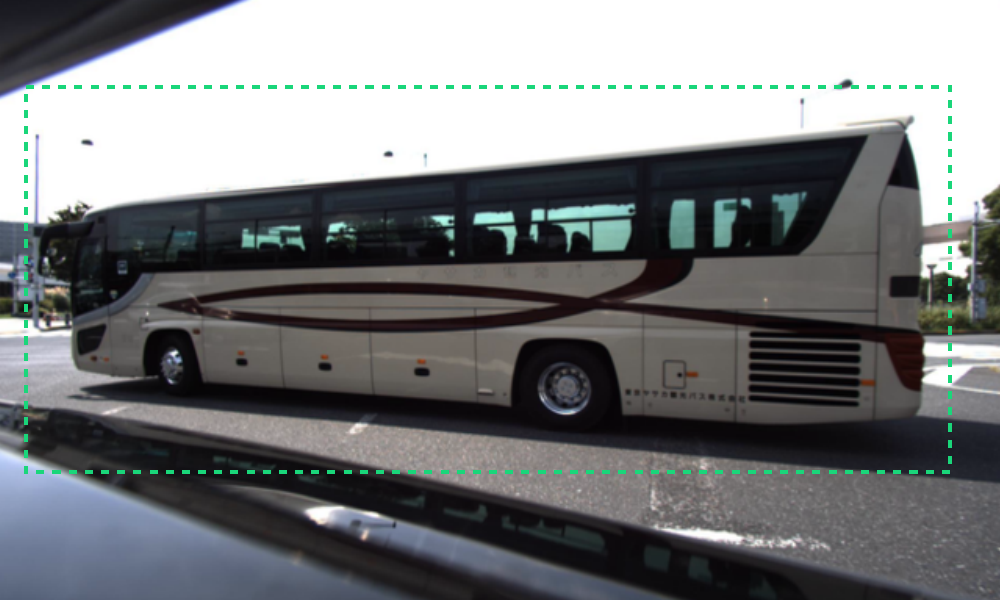} &
\includegraphics[width=0.16\textwidth]{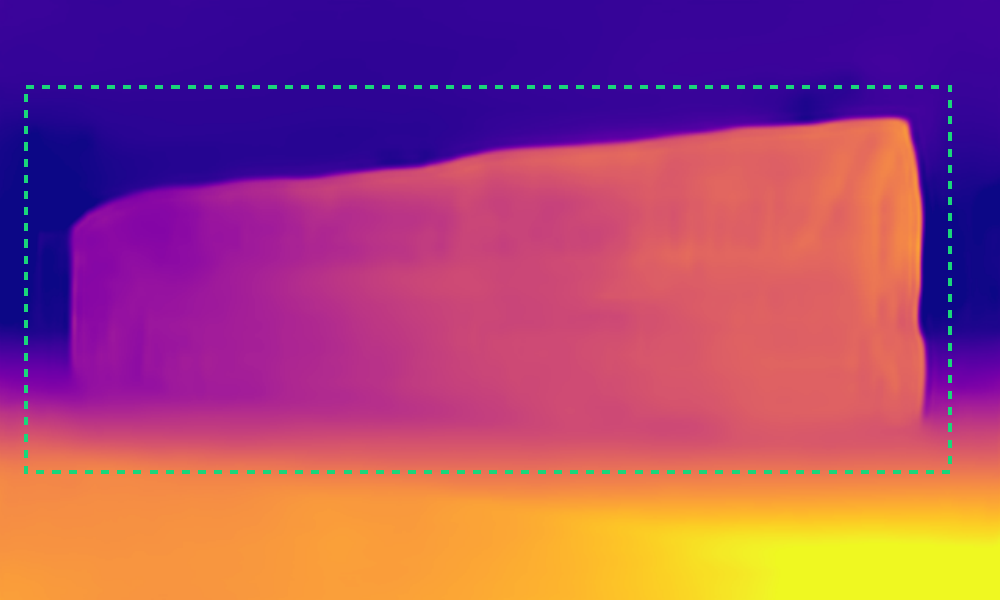} &
\includegraphics[width=0.16\textwidth]{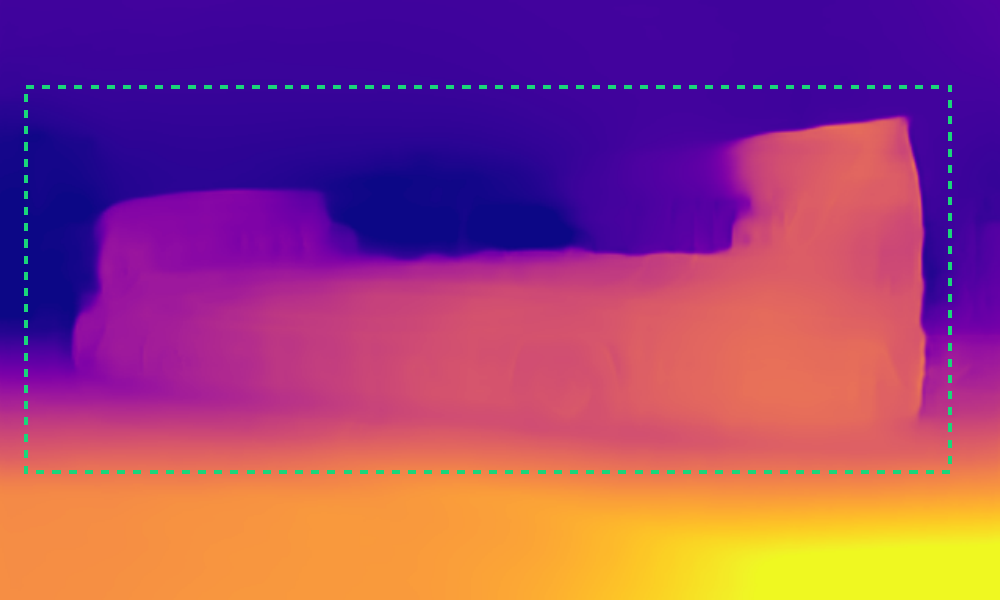} &
\includegraphics[width=0.16\textwidth]{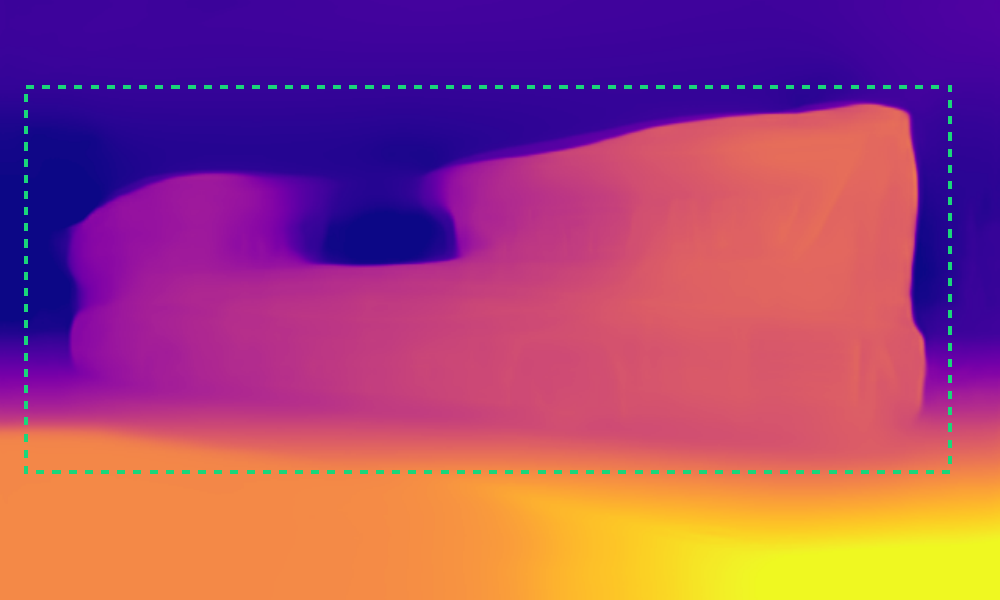} &
\includegraphics[width=0.16\textwidth]{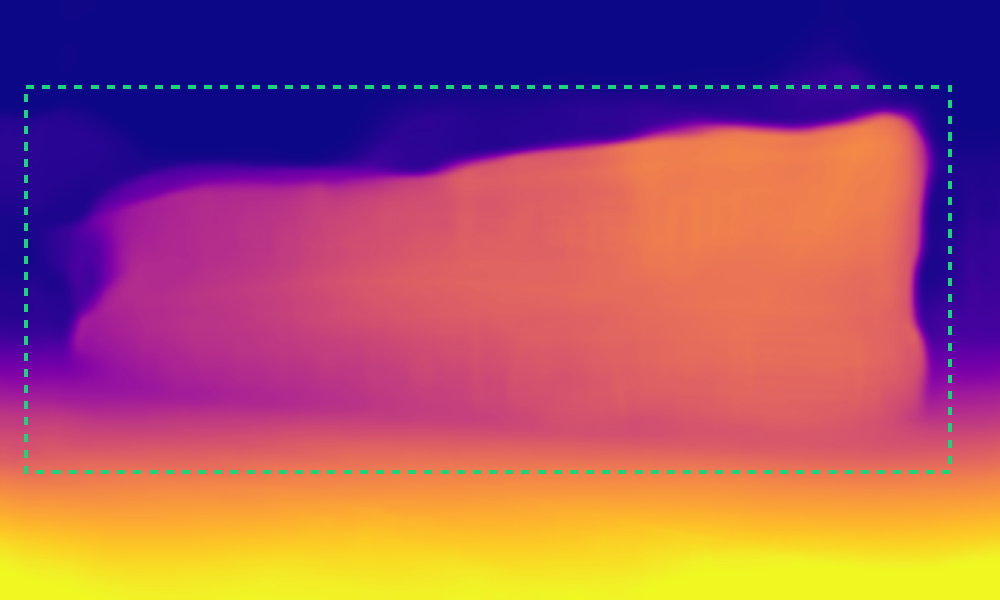} &
\includegraphics[width=0.16\textwidth]{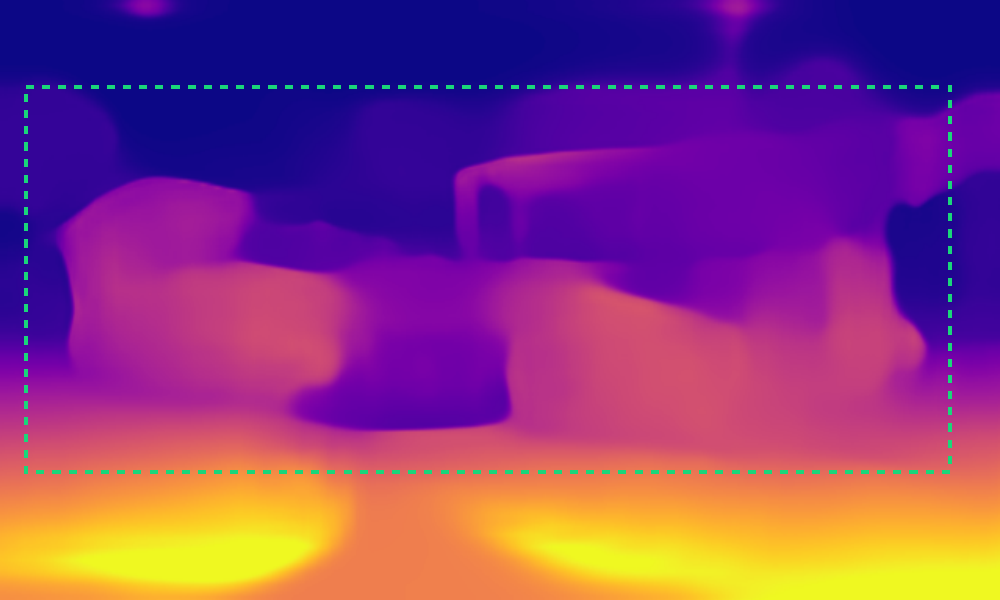} \\
\end{tabular}

\caption{Comparison of depth maps predicted by our method and by state-of-the-art methods on DDAD. Our results show better preserved details and well-defined object boundaries (green bounding boxes). Depth is shown from close in yellow to distant in blue.}
\label{fig:method_comparison}
\end{figure*}

\begin{figure*}[t]
\centering
\begin{subfigure}{0.32\linewidth}
    \centering
    \includegraphics[width=\linewidth]{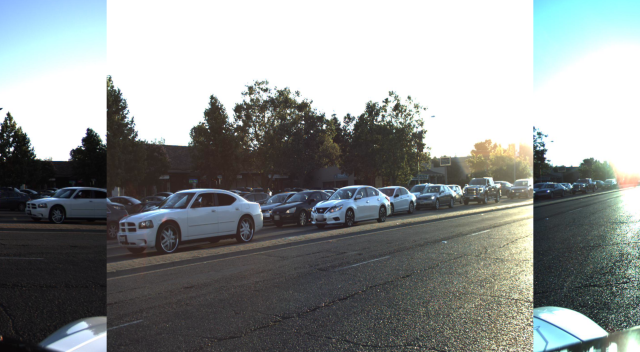}
    \caption{Front-left, front, and front-right image}
\end{subfigure}
\begin{subfigure}{0.32\linewidth}
    \centering
    \includegraphics[width=\linewidth]{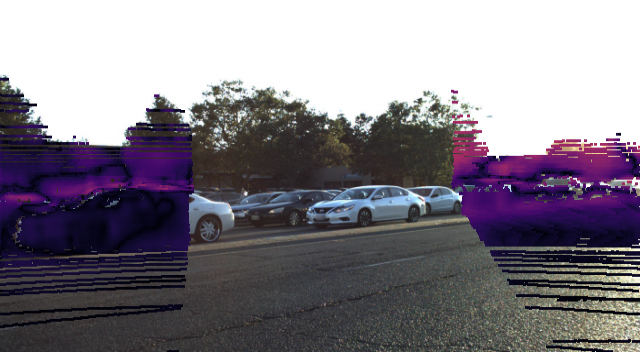}
    \caption{CrossDepth (Ours)}
\end{subfigure}
\begin{subfigure}{0.32\linewidth}
    \centering
    \includegraphics[width=\linewidth]{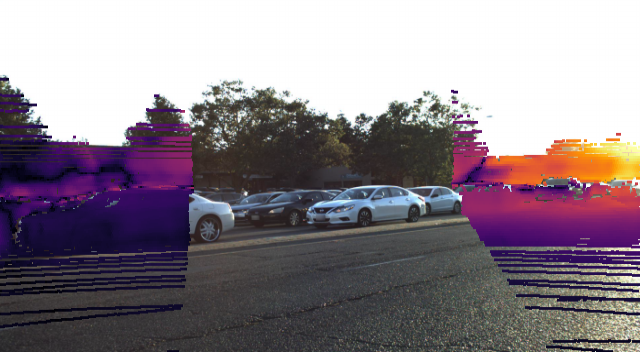}
    \caption{CylinderDepth}
\end{subfigure}

\begin{subfigure}{0.32\linewidth}
    \centering
    \includegraphics[width=\linewidth]{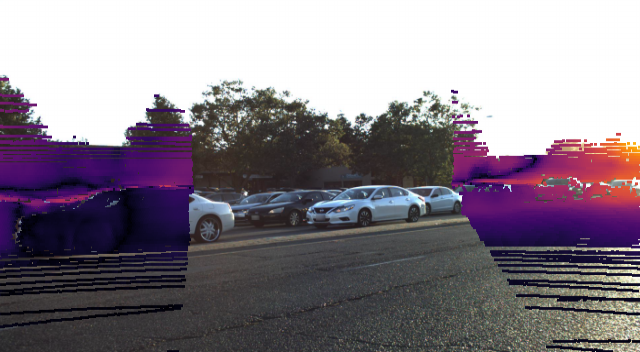}
    \caption{CVCDepth}
\end{subfigure}
\begin{subfigure}{0.32\linewidth}
    \centering
    \includegraphics[width=\linewidth]{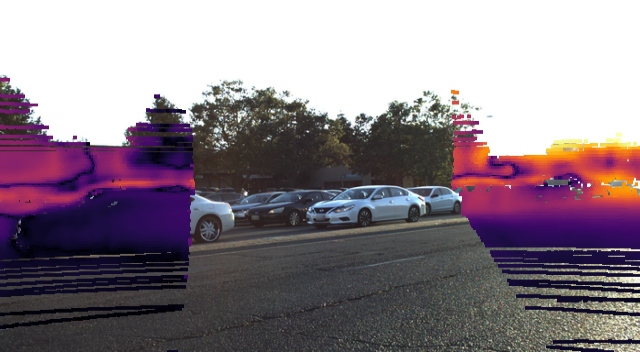}
    \caption{SurroundDepth}
\end{subfigure}
\begin{subfigure}{0.32\linewidth}
    \centering
    \includegraphics[width=\linewidth]{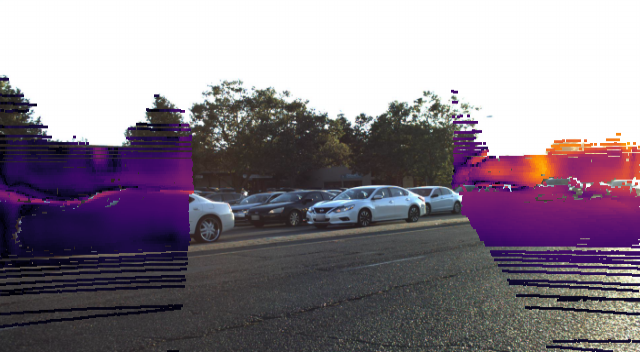}	
    \caption{VFDepth}
\end{subfigure}

\caption{Exemplary depth consistency error maps, computed using the metric described in Sec.~\ref{sec:expsetup} overlayed on the front-image, comparing our approach with state-of-the-art methods on DDAD. Our method maps overlapping regions from the two images to nearby 3D coordinates. In contrast, the other methods exhibit a higher inconsistency in these regions. (a) shows the combined relevant regions from the front-left and front-right images that overlap with the front image. Errors are visualized ranging from black (low error) to yellow (high error).}
\label{fig:3d}
\end{figure*}

\section{Experiments}

\subsection{Experimental Setup}
\label{sec:expsetup}
We compare our method with five state-of-the-art self-supervised approaches: FSM~\cite{guizilini2022full}, SurroundDepth~\cite{wei2023surrounddepth}, VFDepth~\cite{kim2022self}, CVCDepth~\cite{ding2024towards}, and CylinderDepth~\cite{Abualhanud_2026_CVPR}. We report results for the entire image and the overlapping regions under in-domain and cross-domain evaluation. We further conduct ablation studies to assess the effects of cross-image attention, geometric attention constraints, and camera-aware ray embeddings.

\paragraph{Dataset}
Experiments are conducted on two surround-view datasets, DDAD~\cite{guizilini20203d} and nuScenes~\cite{caesar2020nuscenes}. Both datasets provide images from six cameras that jointly cover the full $360^\circ$ surroundings and are used during training and inference, together with LiDAR-based reference depth for evaluation. We use input resolutions of $384\times640$ pixels for DDAD and $352\times640$ pixels for nuScenes. For in-domain evaluation, predictions are evaluated up to 200\,m on DDAD and 80\,m on nuScenes, following the valid ranges of the respective reference depth labels. For cross-domain evaluation on DDAD, we restrict the evaluation range to 80\,m because the model trained on nuScenes is optimized to predict metric depth only up to this range.

\paragraph{Implementation Details}
The depth network uses a pre-trained DINOv3 ViT-S encoder~\cite{simeoni2025dinov3,dosovitskiy2020vit} fine-tuned with LoRA~\cite{hu2021lora}, and a randomly initialized DPT-based depth decoder~\cite{ranftl2021vision}. The pose network uses a ResNet18 encoder initialized with ImageNet-pretrained weights~\cite{he2016deep, russakovsky2015imagenet} and a randomly initialized U-Net decoder~\cite{ronneberger2015u}. Training is performed on four NVIDIA RTX 3090 GPUs with a total batch size of four, each batch comprising $N=6$ surround-view images. The depth and pose networks are optimized jointly using Adam~\cite{kingma2014adam} with $\beta_{1}=0.9$ and $\beta_{2}=0.999$. The initial learning rate is set to $10^{-4}$ and reduced by a factor of $0.1$ using a StepLR scheduler after $\tfrac{3}{4}$ of the total 20 training epochs. For the cross-image attention, we set $B_{\theta}=360$, $B_h=128$, $K_W=5$, and $N_h=8$. For the ray embedding, we set $L_{\mathrm{dir}}=11$. For the loss in Eq.~\ref{Eq:loss}, we set $\lambda_{sp}=0.03$, $\lambda_{spt}=0.1$, $\lambda_{sm}=0.1$, $\lambda_{DCCL}=1\times10^{-3}$, and $\lambda_{MVRCL}=0.2$ based on preliminary experiments.

\paragraph{Evaluation Metrics}
We adopt the standard depth evaluation metrics of~\cite{eigen2014depth}, including Absolute Relative Error (Abs Rel), Squared Relative Error (Sq Rel), RMSE, and the percentage of pixels with an error below a threshold $\delta$. In addition, we report the cross-image Depth Consistency error (Depth Cons) introduced in~\cite{Abualhanud_2026_CVPR}. Depth Cons measures the agreement between predictions of corresponding pixels across images by comparing their Euclidean distances to the origin of the ego coordinate system.

\subsection{Experimental Results}
As shown in Figs.~\ref{fig:method_comparison} and~\ref{fig:3d} and Tab.~\ref{tab:method_compar}, our method achieves higher overall depth accuracy, higher accuracy in overlapping regions, and better cross-view depth consistency than state-of-the-art approaches on most metrics under both, in-domain and cross-domain evaluation. The improvements are particularly pronounced for cross-image consistency in both, the quantitative and qualitative results. 
Compared with CylinderDepth, our method achieves better cross-view consistency while maintaining strong depth accuracy. CylinderDepth aggregates cross-view features using fixed weights determined by cylindrical proximity, whereas our method learns the attention weights within geometrically constrained neighborhoods. In addition, our cross-image attention is applied to all feature layers passed to the decoder rather than only at one feature map scale. Despite applying learned eight-head cross-image attention across feature maps with up to twice the spatial resolution used by CylinderDepth, our method retains a competitive memory footprint, as shown in Tab.~\ref{tab:memory}. The additional memory required during training and inference remains modest relative to the increased complexity of the architecture. Our method is also more memory-efficient than VFDepth, which relies on 3D volumetric feature processing, and SurroundDepth, which applies unconstrained cross-image attention. For more results including an in-depth analysis of remaining limitations refer to the Supp. Material.

\begin{table}[t]
\centering
\setlength{\tabcolsep}{5pt} 
\renewcommand{\arraystretch}{0.95} 
\scriptsize
\resizebox{0.8\columnwidth}{!}{
\begin{tabular}{lcc}
\toprule
Method & Train [GB] & Inference [GB] \\
\midrule
FSM*                & \textbf{5.6} & \textbf{0.5} \\
VFDepth            & 11.0 & 3.3 \\
SurroundDepth      & 12.6 & 1.4 \\
CVCDepth           & \underline{6.0} & \underline{0.7} \\
CylinderDepth & 8.0 & \underline{0.7} \\
\textbf{CrossDepth (ours)} & 8.1 & 1.0 \\
\bottomrule
\end{tabular}%
}
\caption{Efficiency comparison of our method against state-of-the-art in terms of peak allocated memory during training and inference. FSM* denotes the implementation from \cite{kim2022self}. Best results are shown in bold and second-best are underlined.}
\label{tab:memory}
\end{table}

\subsection{Ablation Studies}
\paragraph{Cross-Image Attention}
The ablation results in Tab.~\ref{tab:ablation} show that removing the cross-image attention degrades overall depth accuracy and in particular the cross-image consistency. With unconstrained attention, each token can interact with features from unrelated image regions, including images with no spatial overlap, increasing the likelihood of aggregating irrelevant cross-image features. Explicitly restricting attention to geometrically plausible neighborhoods instead reduces the search space and biases feature aggregation toward likely correspondences. We evaluate multiple neighborhood sizes $K_W$ (see Tab.~\ref{tab:ablationsize}) and observe that relaxing the geometric constraint degrades cross-image depth consistency without improving the overall depth accuracy.

\begin{table}[t]
\centering
\setlength{\tabcolsep}{3pt} 
\renewcommand{\arraystretch}{0.9} 
\footnotesize  
\begin{tabular}{ccccccc}
\toprule
& \multicolumn{4}{c}{\textbf{Overall}} & \multicolumn{2}{c}{\textbf{Overlap}} \\
\cmidrule(lr){2-5} \cmidrule(lr){6-7}
Method & Abs Rel & Sq Rel & RMSE & $\delta < 1.25$ & Abs Rel & Depth Cons \\
\midrule
(a)        & 0.186   & \textbf{2.838} & 11.94  & 74.1 & 0.189 &  5.63  \\
(b)     & 0.193 & 2.944 & 12.53  & 71.8  &  0.198 & 5.39   \\
(c)  & \textbf{0.183}  & \underline{2.869}  &  11.98  & \underline{74.5} & \underline{0.187} & 4.82  \\
(d)     &  0.188  & 2.960 &  11.98 &   73.8 & 0.190  & \underline{4.76}   \\
(e)     &  \underline{0.185}  & 2.894 &  \textbf{11.87} &   \textbf{74.7} & \textbf{0.186}  & \textbf{4.60}   \\
\bottomrule
\end{tabular}
\caption{Ablation study on our method. (a) without cross-image attention; (b) unconstrained attention; (c) frozen backbone weights; (d) focal length scaling; (e) full method: constrained attention, ray embedding and fine tuned backbone. RMSE, Sq Rel and Depth Cons are given in [m]. $\delta$ is given in [\%]. Abs Rel is unit-free. Results are reported for the entire images and for overlapping regions on the DDAD dataset. Best results are shown in bold and second-best are underlined.}
\label{tab:ablation}
\end{table}

\begin{table}[t]

\centering

\setlength{\tabcolsep}{6pt}
\renewcommand{\arraystretch}{0.9}
\footnotesize

\begin{tabular}{ccc}

\toprule

$K_W$ & Overall (Abs Rel) & Overlap (Depth Cons) [m]\\

\midrule

$5$ & \textbf{0.185} & \textbf{4.60} \\
$10$ & \underline{0.187} & \underline{5.14} \\
$20$ & \underline{0.187} & 5.55 \\

\bottomrule

\end{tabular}

\caption{Ablation study on the neighborhood size of the constrained cross-image attention. Metrics are reported for overall depth accuracy and cross-image consistency within overlapping regions on DDAD. Abs Rel is unit-free. Best results are shown in bold and second-best are underlined.}

\label{tab:ablationsize}

\end{table}

\paragraph{Ray Embedding}
Incorporating camera intrinsics through per-token ray embeddings further improves depth accuracy (see Tab.~\ref{tab:ablation}). In contrast, focal-length normalization compensates for changes in camera intrinsics using a single global multiplicative scale. It does not explicitly account for how the projection geometry varies across pixels. Ray embeddings instead provide the model with the viewing direction of each token, allowing it to learn a spatially varying and non-linear relationship between monocular appearance cues, camera intrinsics, and metric depth.

\paragraph{Encoder Backbone}
We also evaluate a variant with the backbone kept fully frozen. Allowing lightweight LoRA fine-tuning yields a modest improvement, indicating that adapting the pretrained representation to the depth estimation task is beneficial. However, the relatively small gap to the frozen-backbone variant suggests that the pretrained DINO features already encode much of the visual information required for monocular depth estimation. This observation is consistent with the findings of~\cite{danier2025depthcues,simeoni2025dinov3}.
\section{Conclusion}
In this work, we introduce a unified framework for self-supervised metric depth estimation from surround-view camera rigs. Our method addresses two main sources of cross-view inconsistency: differences in the visual context of each image, handled through geometry-constrained cross-image attention, and variations in monocular appearance cues caused by different camera intrinsics, addressed using per-token camera-ray embeddings. We find that geometrically constraining cross-image attention positively affects depth accuracy and cross-image consistency as opposed to unconstrained attention. However, several limitations remain. Because cross-image attention is applied only to the encoder features and not within the decoder, localized depth artifacts can appear near object boundaries. Finally, due to computational constraints, we have not evaluated the method at substantially larger training scales, leaving its scalability to larger and more diverse datasets for future investigation.
{
    \small
    \bibliographystyle{ieeenat_fullname}
    \bibliography{main}
}

\end{document}